\documentclass[conference]{IEEEtran}
\usepackage{cite}
\usepackage{amsmath,amssymb,amsfonts}
\usepackage{algorithm}
\usepackage{algorithmic}
\usepackage{graphicx}
\usepackage{textcomp}
\usepackage{xcolor}
\def\BibTeX{{\rm B\kern-.05em{\sc i\kern-.025em b}\kern-.08em
    T\kern-.1667em\lower.7ex\hbox{E}\kern-.125emX}}

\ifCLASSOPTIONcompsoc
\usepackage[caption=false,font=normalsize,labelfont=sf,textfont=sf]{subfig}
\else
\usepackage[caption=false,font=footnotesize]{subfig}
\fi

\DeclareMathOperator*{\argmax}{argmax}

\begin{document}

% no suggestion about quality metric, just about the final objective... is this okay?
%\title{Towards Physics from vision for task-oriented grasping}
%\title{Towards Physics Perception for \\ Task Oriented Grasping}
%\title{Towards Task Oriented Grasp Quality Metrics \\ \LARGE{-- Affordance Functions Prediction from Vision --}}
\title{Towards Affordance Prediction with Vision via \\ Task Oriented Grasp Quality Metrics }
% nelle guidelines di IEEE dicono:
% Note: Sub-titles are not captured in Xplore and should not be use
% ...è un problema?

\author{\IEEEauthorblockN{Luca Cavalli}
\IEEEauthorblockA{
\textit{Politecnico di Milano}\\
Milan, Italy \\
luca3.cavalli@mail.polimi.it}
\and
\IEEEauthorblockN{Gianpaolo Di Pietro}
\IEEEauthorblockA{
\textit{Politecnico di Milano}\\
Milan, Italy \\
gianpaolo.dipietro@mail.polimi.it}
\and
\IEEEauthorblockN{Matteo Matteucci}
\IEEEauthorblockA{
\textit{Politecnico di Milano}\\
Milan, Italy \\
matteo.matteucci@polimi.it}
}

\newcommand{\tmp}[1]{\color{red}#1 \par \color{black}}

\maketitle

\begin{abstract}
%Robotic manipulation of novel objects involves complex physical interactions whose exact planning is prevented by the uncertainty in perception.
%In this paper a novel approach to overcome this limitation in task-oriented grasping is proposed.
%Quantification of the quality of results is a key element towards solid results.
While many quality metrics exist to evaluate the quality of a grasp by itself, no clear quantification of the quality of a grasp relatively to the task the grasp is used for has been defined yet. In this paper we propose a framework to extend the concept of grasp quality metric to task-oriented grasping by defining affordance functions via basic grasp metrics for an open set of task affordances. We evaluate both the effectivity of the proposed task oriented metrics and their practical applicability by learning to infer them from vision. Indeed, we assess the validity of our novel framework both in the context of perfect information, i.e., known object model, and in the partial information context, i.e., inferring task oriented metrics from vision, underlining advantages and limitations of both situations. In the former, physical metrics of grasp hypotheses on an object are defined and computed in known object model simulation, in the latter deep models are trained to infer such properties from partial information in the form of synthesized range images. 
%Task related affordances are explicitly considered as functions of these properties and new and effective grasps are obtained by optimizing the task affordance functions.
%Task-related fitness functions based on the inferred properties allow to plan for a suitable grasp for the given task by optimization of the fitness. 
\end{abstract}


%\tmp{General TODOs:}
%\tmp{refine the images: photoshop axes away and highlight the point of use in red to make it clearly visible when relevant} -> done!

\section{Introduction}
The research community has spent much effort in tackling the problem of grasping novel objects in different settings~\cite{BlindGraspCorr}~\cite{2DBigDataGrasp}~\cite{CameraInHand}~\cite{ECVFeatures}~\cite{AmazonChallenge2017Princeton} with the objective of holding objects robustly with robotic manipulators; however, real manipulation tasks go far beyond holding the objects and the quality of a grasp depends on the task it is meant to support. While many quality metrics exist to evaluate the quality of a grasp by itself~\cite{roa2015grasp}~\cite{miller1999examples}, no clear quantification of the quality of a grasp relatively to a task has been defined. In this paper we propose a framework to extend the concept of quality metric to task-oriented grasping by defining general physical measures for an open set of task affordances. We evaluate both the results provided by such metrics and their applicability in practice by learning to infer them from vision. 

% These quality metrics, in the form of functions of the geometrical and physical properties of an object and a grasp, directly encode and quantify the affordance of an object with respect to a task relatively to a fixed manipulator.

\begin{figure}
    \centering
    \includegraphics[width=0.9\linewidth]{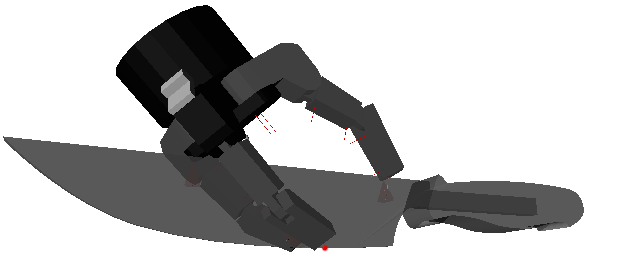}
    \caption{Best grasp, according to the proposed metrics, relatively to the affordance function $\tilde{F}_{cut}$ defined in Section~\ref{subsec:aff_func_def} for the cutting task using a common kitchen knife}
    \label{fig:intro_img}
\end{figure}

More formally, given a grasp $G$ on an object $O$ and a point $U$ on the surface of $O$ (which is the point where we plan to use the object, when the task requires one), we define the affordance function $F_T : (O, G, U) \mapsto \mathbb{R}$ to define the affordance of any possible grasp $G$ and use hypothesis $U$ with respect to task $T$. The final objective is to optimize for the best grasp, object and use location $(O, G, U)$ that maximizes a given affordance function, as shown in Figure~\ref{fig:intro_img}.
We are interested in finding a set of metrics, as functions of the triplet $(O, G, U)$, to encode significant static and geometric properties of the $(O, G, U)$ system itself. Examples of such metrics are the local geometry of $O$ around $U$, or the minimum sum of contact forces needed to hold the object $O$ with grasp $G$ under a given gravity vector. A complete description of the metrics used in this work, not to be considered as an exhaustive list, is reported in Section~\ref{subsec:metrics}. 
%\tmp{Qui ci starebbe un paragrafo di contesto in cui spieghiamo che quel che ci interessa e' essere in grado di prevedere il miglior uso possibile di un oggetto senza necessariamente riconoscerlo, ma solo dalla sua forma geometrica e dalle sue proprieta' fisiche. Questo e' un po' quello che facciamo noi quando pensiamo in modo creativo agli usi possibili degli oggetti ... insomma una storiella catchy per "vendere l'idea"}

The proposed approach allows the inference of the affordance of objects without having it bound to their semantic category: semantic information on objects defines their standard use meant for humans, which is not necessarily the only nor even optimal use for robots. Semantics greatly simplifies the task of affordance perception, but it gives no guarantee of optimality, particularly with robot actuators which differ substantially from human hands and arms. Take, for instance, the classical human grasp of a hammer with the wrist direction parallel to the beating direction: actuating such grasp with a Barrett hand~\cite{BarrettHand} which has only a rotational degree of freedom on the wrist would have the same efficacy in beating as a human with a locked wrist, even without taking into consideration the decreased number of fingers and much reduced tangential and torsional friction in contacts.
%We approximate $F_T$ with the use of elementary metrics $f: (O, G, U) \mapsto \mathbb{R}^n$ to $\tilde{F_T} : \mathbb{R}^n \mapsto \mathbb{R}$
%To allow the inference of elementary metrics from vision, under incomplete object geometry, we decouple the encoding of grasp and use locations from the object geometry through the use of a simple but effective fixed grasp policy as shown in Figure~\ref{fig:grasp_policy}. 
%\tmp{Qui sembra un po' tronco, nella rilettura andra' completato un po' ...}
% tagliato per risparmiare spazio che scarseggia, in fin dei conti è un dettaglio che viene spiegato ampiamente dopo

To validate our framework, we have collected a dataset of grasps and computed elementary graps metrics by using the GraspIt! simulator~\cite{GraspIt!} on the Princeton Shape Benchmark object models~\cite{shilane2004princeton}. Then we have trained deep models to infer those elementary metrics from range images taken in a simulated camera-in-hand setting to assess the applicability of our framework to more realistic partial-information settings. 

In our contribution, we define a framework for quality assessment of task-oriented grasps, we qualitatively validate such framework, we provide a GraspIt! plugin to produce labelled data with minimal to no human intervention, thus in an extremely scalable way, and we generated a dataset of 400M evaluated grasps on 22 objects of the Princeton Shape Benchmark which we plan to make public in the near future. Moreover, we propose and benchmark models to tackle the problem of learning to infer such metrics from vision. Preliminary results shows direct optimization of affordance functions in simulation produces new and creative grasps which fit the specific actuator in use for the selected task, while direct inference of such metrics from vision is yet an open challenge and there are great margins for further improvement.

% \tmp{Add here paper outline} -> missing space!

\section{Related works}
Many researchers have worked towards the understanding and formalization of the concept of affordances~\cite{gibson1966senses}~\cite{Michaels}~\cite{csahin2007afford}; they have been inspiring for roboticists to work within the affordances framework to define the autonomous interaction of a robot with an unknown environment. In our work we investigate the broad category of robot affordances focusing on the specific application of task-oriented grasping.
Within this context, one of the first approaches towards task-oriented grasping, reported in~\cite{PreshapesTaskGrasp2007}, proposed to encode the task in physical terms (e.g., applying a momentum on a handle to open a door) and then to solve the problem of grasp planning by hardcoding hand postures and their association with tasks; the method has shown good performance in the expected domain, but poor generalization capabilities. Later works formalized the problem via graphical models, distinguishing task, object features, action features and constraint features. In particular, authors of~\cite{song2010learning} proposed the use of such formalization and they have been able to effectively learn to infer the likelihood of grasp approach directions with respect to a human-labelled ground truth. The main limitation of this work, in our opinion, is the human intervention, which makes the real definition of the tasks implicit and prevents the scalability of the dataset that can be generated for learning without tedious human teaching. The direct intervention of human judgment on semantics to evaluate the quality of grasp hypotheses with respect to a given task is nevertheless a common approach to many research works, like~\cite{SemanticGrasping2012} and~\cite{SemanticGrasping2014} in which authors prove the effectiveness of a human-labelled semantic approach with real robot manipulations. More recently,~\cite{detry2017task} has proposed to label mesh vertices in simulated objects as being graspable or not according to some task, so that many scene examples can be produced and automatically labeled via simulation. This allowed the system to automatically segment graspable and not graspable regions of objects in cluttered scenes, but still the expressivity of this method is restricted to specifying graspable or not graspable surfaces. Reference~\cite{Lakani2019} proposed a bottom-up approach for affordance perception by object parts which detects the local geometry of patches of the object and provides pixel-level affordance segmentation for pre-defined affordance classes. Their method is again based on a dataset~\cite{myers2015affordance} of 10000 pixel-labelled RGB-D images which have been hand labelled.

A common limitation of these approaches is the vague definition of task affordances which passes through the human labeling of the ground truth. This entails a great limitation in the size of the data that can be produced for learning and poses questions about the optimality and validity of the labels with respect to the actual task performance with a different actuator than the human hand.

% \tmp{visto che qui ne fate un punto chiave, nei risultati vanno citate o stressate isa la feasibility sia la scalability}

%%%%%%%%%%%%%%%%%%%%%%%%%%%%%%%%%%%%%%%%%%%%%%%%%%%%%%%%%%%%%%%%%%%%%%%%%%%%%%%%%%%%
%%%%%%%%%%%%%%%%%%%%%%%%%%%%%%%%%%%% PROPOSED APPROACH %%%%%%%%%%%%%%%%%%%%%%%%%%%%%
%%%%%%%%%%%%%%%%%%%%%%%%%%%%%%%%%%%%%%%%%%%%%%%%%%%%%%%%%%%%%%%%%%%%%%%%%%%%%%%%%%%%

 \begin{figure*}
 \begin{center}
%    \centering
\hfill
  \subfloat[\label{fig:grasp_policy:a}]{%
       \includegraphics[width=0.3\linewidth,trim={40 30 0 20},clip]{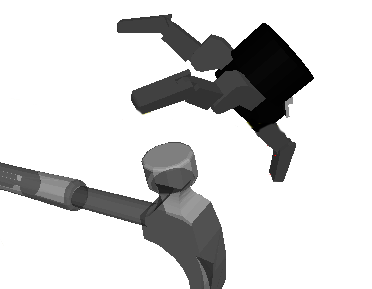}}
%    \hfill
  \subfloat[\label{fig:grasp_policy:b}]{%
        \includegraphics[width=0.3\linewidth,trim={0 0 0 50},clip]{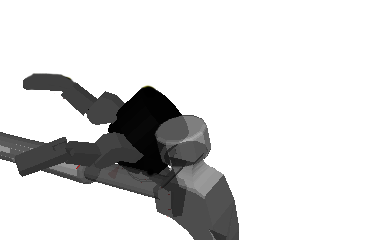}}
%    \hfill
  \subfloat[\label{fig:grasp_policy:c}]{%
        \includegraphics[width=0.3\linewidth,trim={0 0 0 50},clip]{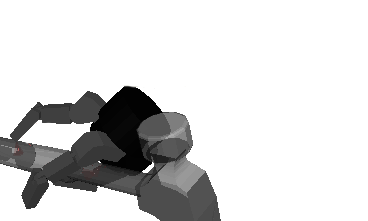}}
        \end{center}
  \caption{The three phases of our grasp policy: (a) the pregrasp parameters determine the initial position and posture of the hand (b) the hand approaches in a straight line until a contact is made (c) fingers close until they make contact or they are completely closed}
  \label{fig:grasp_policy} 
\end{figure*}

\section{Proposed approach}
%\subsection{Problem formulation}
\label{subsec:problem}
% definition of basic sets
Let $\mathcal{O}$ be the set of possible object surfaces with friction and softness properties defined at each point, let $\mathcal{G}(O)$ defined on an object $O \in \mathcal{O}$ be the set of possible grasps determined by the hand embodiment, degrees of freedom, contact locations on the object and contact nature (e.g., frictionless, hard contact or soft contact), and $\mathcal{U}(O)$ be the set of points on the surface of the same object that can be considered as points of use. 

% definition of affordance function
Let $O \in \mathcal{O}$, $G \in \mathcal{G}(O)$, $U \in \mathcal{U}(O)$ , then we define the \textit{affordance function} of task $T$ as $F_T(O, G, U) \mapsto \mathbb{R}$ such that $F_T(O_1, G_1, U_1) > F_T(O_2, G_2, U_2)$ if and only if the grasp and use hypothesis $(O_1, G_1, U_1)$ is more suitable than the hypothesis $(O_2, G_2, U_2)$ for task $T$, thus defining an affordance ordering of an object grasp for task $T$.

% definition of elementary static and geometrical metrics as functions of grasp, object, use
As we want a compact representation of the affordance function, we approximate $F_T$ as a $\tilde{F}_T : \mathbb{R}^n \mapsto \mathbb{R}$ by mapping the triplet $(O, G, U)$ into a metric vector $\phi \in \mathbb{R}^n$ through a function $\Phi(O, G, U) \mapsto \mathbb{R}^n$. This metric vector is a collection of metrics encoding the geometrical and static physical properties of the triplet $(O, G, U)$ which are relevant to approximate $F_T$. In Section~\ref{sec:metrics} we provide some examples of basic metrics $\phi$ and examples on how they could be used to hardcode $\tilde{F}_T$ for some reference tasks.

\subsection{Achieving Object Semantics Independence}
The complete object geometry is generally not available in real world applications, in particular when our long term goal is to infer object affordance from vision with no hardwired semantics. To achieve such goal, we need to frame the problem in the context of uncertain and incomplete information about the object by decoupling the grasp and use location description from the exact object geometry and possibly its semantics.

%\paragraph{Decoupling grasps} 
Recall here that the complete description of a grasp requires the geometry and nature of contact points on the grasped object, and the grasp itself needs to be actuated by a grasping policy. If we assume the grasping policy to be deterministic, then we can define it as a function $GP(p_0, O) \mapsto \mathcal{G}(O)$ that maps an initial state $p_0\in\mathcal{P}_0$ and an object $O$ into the final grasp $G \in \mathcal{G}(O)$. To decouple from the specific grasp, and its parameters, we fix a grasping policy that allows a sufficient exploration of the grasps space $\mathcal{G}$ via the space $\mathcal{P}_0$ of possible initial states, which we call \textit{pregrasps}. 

In particular, we select a simple, but effective, grasping policy defined as follows: from an initial position of the hand with open fingers, we advance towards a fixed direction until the first contact is made, then the fingers are closed until all of them either make contact or are completely closed. 
Jointly with the use of eigengrasps~\cite{ciocarlie2007dexterous} to describe the degrees of freedom of the hand, this policy allows for a further reduction in the dimensionality of the problem for the translational degree of freedom saved in describing the position of the hand in the space, as approaching the object in a straight line makes the approach direction invariant with respect to the policy. If the eigengrasp parameters are divided into $h$ parameters used to close the grasp and $k$ parameters used to set the initial hand posture, then we can consider $p_0 \in \mathbb{R}^{5+k}$, where the 5 extra parameters describe the initial position and rotation of the hand. The $h$ eigengrasp parameters used to close the hand are fixed to an open position on the pregrasp, thus they do not contribute to the dimensionality of the pregrasp itself.

%\paragraph{Decoupling use locations} 
On the other hand, use locations are points on the bidimensional surface of each specific object; we decouple them from the specific object by defining use directions and a direction mapping function $DM(d, O) \mapsto \mathcal{U}(O)$ in complete analogy to the grasp decoupling solution. Our direction mapping assumes $d \in \mathbb{R}^2$ to be the spherical coordinates of a directed ray centered in the center of mass of object $O$ and output in the farthest point $U \in \mathcal{U}(O)$ which is the intersection of such ray with the outer object surface.

\subsection{Metrics Inference from Vision}
\label{subsec:inference}
As our goal is to make robots able to use unknown objects in a task consistent way, we need the robot to be able to perceive their affordances via sensors. In particular, we focus on vision being it an extremely common and effective tool to take information from the environment in real applications. We define our inference setting by focusing on the specific case of single range images taken from camera-in-hand perspective: such case provides only local geometry information about the object around the expected location of the planned grasp.
We want to learn a model that predicts the elementary metrics $\phi \in \mathbb{R}^n$ of a triplet $(O, G, U)$ with only partial information about the observed object $O$. Indeed, predicting the vector $\phi$ would allow to estimate the affordance function $F_T$ for any task $T$ for which we can define an $\tilde{F}_T$.

Let $D: \mathcal{P}_0 \times \mathcal{O} \mapsto \mathbb{R}^{h \times w}$ be the function that maps a pregrasp $p_0$ on an object $O$ to the depthmap of size $h \times w$ from the camera-in-hand perspective of $p_0$. Then we want to learn a model $\mathcal{M}^\Phi$ that approximates the mapping from $(p_0, D(p_0, O), d)$ to $\Phi(O, GP(p_0, O), DM(d, O))$ where $O$ is an object, $p_0$ is a pregrasp, $d$ is a use direction, and $\Phi$ is the metric extraction function, under the grasping policy $GP$ and the direction mapping function $DM$.

We assume to be able to learn model $\mathcal{M}^\Phi$ from a dataset of tuples $(O, p_0, d, \Phi(O, GP(p_0, O), DM(d, O))$ obtained via uniform sampling of $p_0$ and $d$ values on a set of available objects models and computing the true values of $\Phi(O, GP(p_0, O), DM(d, O))$ via simulation. Details on our data collection setup are explained in Section~\ref{subsec:data}. 

%\subsection{Model definition}
To structure the learning task, we define the model $\mathcal{M}^\Phi$ as the composition of two models: an input value $(p_0, D(p_0, O), d)$ is first classified by a binary classifier $\mathcal{M}^\Phi_C$ to infer whether it represents a ``good'' grasp worth further evaluation or not. We define ``good'' grasps those respecting a minimum quality independently from the task, thus employing state of the art grasp quality metrics to generate the ground truth. The samples classified as positive then pass through a regression model $\mathcal{M}^\Phi_R$ that infers the metrics $\phi$ with the implicit assumption that the grasp is indeed a quality grasp.

For both models $\mathcal{M}^\Phi_C$ and $\mathcal{M}^\Phi_R$ we propose and evaluate the two architectures of the convolutional neural network (CNN) and the PointNet architecture~\cite{qi2017pointnet}. Both architectures encode the available geometry information (in the form of range image for the CNN or as the equivalent projected point cloud for the PointNet) in a feature vector, we then apply late-fusion of the other input parameters $p_0$ and $d$ on this feature vector and output the classification label or the inferred regression value with a classical fully connected network.

%%%%%%%%%%%%%%%%%%%%%%%%%%%%%%%%%%%%%%%%%%%%%%%%%%%%%%%%%%%%%%%%%%%%%%%%%%%
%%%%%%%%%%%%%%%%%%%%%%%%%%%%%% FEATURE SELECTION %%%%%%%%%%%%%%%%%%%%%%%%%%
%%%%%%%%%%%%%%%%%%%%%%%%%%%%%%%%%%%%%%%%%%%%%%%%%%%%%%%%%%%%%%%%%%%%%%%%%%%

\section{Task Oriented Grasp Metrics}
\label{sec:metrics}
In this work we concentrate on the sample tasks of beating, cutting and picking, which are defined by their $\tilde{F}_{beat}$, $\tilde{F}_{cut}$ and $\tilde{F}_{pick}$ in Section~\ref{subsec:aff_func_def}. Such tasks have been selected with the idea of the kitchen assistant robot in mind, considering some very different tasks that may happen to be requested in this sample application. We first define a set of grasp metrics, then define from these the corresponding affordances.

\subsection{Basic Grasp Metrics}
\label{subsec:metrics}
We consider the following set of elementary metrics of $(O, G, U)$ which should not be considered as exhaustive:

\newcommand{\rob}{\epsilon}
\newcommand{\inertia}{I}
\newcommand{\handEffortImpact}{E_i}
\newcommand{\handEffortHold}{E_h}
\newcommand{\dischargeEfficiency}{\delta}
\newcommand{\pointForce}{U_\tau}
\newcommand{\useGeom}{{U_g}}
% \newcommand{\usePhys}{U_p}

 %   \item \textbf{Mass ($\mass \in \mathbb{R}$)}: the mass of object $O$
    \paragraph{Grasp robustness ($\rob \in \mathbb{R}$)} is a real number describing the robustness of the grasp. We use the Epsilon metric described in~\cite{miller1999examples} as a builtin in the GraspIt! simulator~\cite{GraspIt!}. Force closure grasps have $\rob > 0$ and higher robustness implies a greater minimum perturbance is needed to break the grasp.
    \paragraph{Rotational inertia ($\inertia \in \mathbb{R}$)} quantifies the rotational inertia around the axis of rotation of the wrist of the hand assuming a unitary density of the object and assuming the hand to be integral with the whole object. It does not take into account the mass of the hand itself.
    \paragraph{Hand effort on impact ($\handEffortImpact \in \mathbb{R}$)} describes the effort of the hand to balance the impact forces after a rotation around the wrist. It assumes a fixed average inertial torque in a small $\Delta t$ during the impact which is directly proportional to $\inertia$ and a free contact force on the use location towards the normal direction. This metric takes the value of the minimum sum of the contact forces of the hand constrained to the contact friction cones to balance the inertial torque, $\infty$ if the minimization problem is unfeasible.
    \paragraph{Hand effort on hold ($\handEffortHold \in \mathbb{R}^6$)} is a vector of six independent values which quantify the hand effort to balance a different gravity vector. The hand effort is the minimum sum of all contact forces constrained to the contact friction cones that balance a given unitary force of gravity, $\infty$ if such problem is unfeasible. The six gravity vectors chosen are aligned with the three coordinate axes (once in the same direction, once opposite) of the object mesh as all meshes that we used in the Princeton Shape Benchmark have been designed by humans that gave a semantic meaning to the coordinate axes directions.
    \paragraph{Momentum discharge efficiency ($\dischargeEfficiency \in \mathbb{R}$)} quantifies the efficiency of discharging the rotational inertia of the wrist on the object use location. It quantifies the alignment between the inertial torque and the torque generated by a force aligned with the use location normal vector towards the inside of the object surface. It is computed as the dot product of the two normalized vectors, clipped to zero in case of negative values.
    \paragraph{Force transmitted to use ($\pointForce \in \mathbb{R}$)} quantifies the force that can be transmitted to the use location using constrained contact forces. It assumes all contact forces are constrained by their friction cones and have unitary maximum normal forces. It takes the value of the maximum force on the use location towards the use location normal guaranteeing static conditions.
    \paragraph{Use local geometry ($\useGeom \in \mathbb{R}$)} describes how much the use location has the shape of an edge. It is obtained by fitting a quadratic function on the vertices of the triangles near the use location (including all the triangles that share at least one vertex with the triangle where the use location lies) and extracting the eigenvalues of the hessian matrix of such quadratic function. The two eigenvalues $\lambda_1$ and $\lambda_2$ are the two principal component curvatures, so we quantify an edge with the expression $(\lambda_1-\lambda_2)^2$ to identify locations with a great difference in local curvatures.
%    \item \textbf{Use local properties ($\usePhys \in \mathbb{R}^2$)}: <placeholder for friction and hardness>

\subsection{Affordance Functions from Basic Metrics}
\label{subsec:aff_func_def}

On top of these metrics we define the affordance functions for $T\in\{beat, cut, pick\}$. In this preliminary study affordance functions have been designed by hand, to validate the feasibility of the framework, in future works we aim to learn them by optimizing task execution efficacy. 

\paragraph{Beating} The classical beating action of a hammer with a human hand requires dexterous movements of the wrist which would need moving the whole robotic arm to be reproduced on a Barrett hand. For this reason we assume that the beating action will be executed by the robotic actuator by simply rotating the hand clockwise around the wrist. We require that the hold is stable over a minimum threshold and that the rotational energy gets discharged almost entirely on the point of use. We want to maximize the ratio of the energy that we can incorporate into the rotation (assuming a maximum rotational speed) over the actual hand effort of keeping the object stable on the impact.

% beating
 \begin{algorithmic}[1]
 \renewcommand{\algorithmicrequire}{\textbf{Input:}}
 \renewcommand{\algorithmicensure}{\textbf{Output:}}
 \REQUIRE $\rob, \dischargeEfficiency, \inertia, \handEffortImpact, \handEffortHold$
 \ENSURE $\tilde{F}_{beat}$
 \IF {$\rob < \tau_\rob $ OR $ \dischargeEfficiency < \tau_\dischargeEfficiency$ OR $\sum_{i=1}^{6} \handEffortHold[i] == \infty$}
 % \tau_\rob = 0.3, \tau_\dischargeEfficiency = 0.95
    \RETURN $-\infty$
 \ELSE
    \RETURN $\frac{\inertia}{\handEffortImpact}$
 \ENDIF
 \end{algorithmic}

 \paragraph{Cutting} The action of cutting is extremely complex by itself and varies greatly with different materials and their surface and micro-structural properties. A complete physical study of this particular task is not our objective; we simplify it considering as approximation that greater force provides cuts if executed on a thin enough edge. %Consequently we obtain the following $\tilde{F}_{cut}$:
 % cutting
  \begin{algorithmic}[1]
 \renewcommand{\algorithmicrequire}{\textbf{Input:}}
 \renewcommand{\algorithmicensure}{\textbf{Output:}}
 \REQUIRE $\rob, \pointForce, \useGeom$
 \ENSURE $\tilde{F}_{cut}$
 \IF {$\rob < \tau_\rob $ OR $ \useGeom < \tau_\useGeom$}
 % \tau_\rob = 0.3, \tau_\useGeom = 10
    \RETURN $-\infty$
 \ELSE
    \RETURN $\pointForce$
 \ENDIF
 \end{algorithmic}

 \paragraph{Picking} Picking an object (as the first part of the pick-and-place task) only strictly requires a stable grasp for a successful pick. However, different stable grasps may imply very different effort from the hand actuator to balance the force of gravity on the object. For this reason we require a stable grasp and minimize the sum of the contact forces required to balance the force of gravity in the six directions evaluated by the $\handEffortHold$ metric. Notice that an unstable grasp will need to have at least one evaluated direction of gravity that the grasp cannot hold, thus we do not check the $\rob$ metric.
 % picking
  \begin{algorithmic}[1]
 \renewcommand{\algorithmicrequire}{\textbf{Input:}}
 \renewcommand{\algorithmicensure}{\textbf{Output:}}
 \REQUIRE $\handEffortHold$
 \ENSURE $\tilde{F}_{pick}$
    \RETURN $-\sum_{i=1}^{6} \handEffortHold[i]$
 \end{algorithmic}

%%%%%%%%%%%%%%%%%%%%%%%%%%%%%%%%%%%%%%%%%%%%%%%%%%%%%%%%%%%%%%%%%%%%%%%%%%%%%
%%%%%%%%%%%%%%%%%%%%%%%%%%%%%% EXPERIMENTAL RESULTS %%%%%%%%%%%%%%%%%%%%%%%%%
%%%%%%%%%%%%%%%%%%%%%%%%%%%%%%%%%%%%%%%%%%%%%%%%%%%%%%%%%%%%%%%%%%%%%%%%%%%%%

\section{Framework Validation}

To assess the feasibility of the proposed approach we performed a set of experiments focused on the tasks of beating, cutting and picking; this choice has driven the selection of basics metrics to encode in function $\Phi$ and the definition of functions $\tilde{F}_{beat}$, $\tilde{F}_{cut}$ and $\tilde{F}_{pick}$ in the previous section. In the validation we aim at: 
\begin{enumerate}
    \item \textbf{Validating the framework} by showing that $\argmax_{G, U} \tilde{F}_T(\Phi(O, G, U))$ for some selected tasks $T$ provides grasps and use locations that are semantically meaningful with respect to the semantics of task $T\in\{beat, pick, cut\}$
    \item \textbf{Assessing the feasibility of learning} a model $\mathcal{M}^\Phi$ that can infer basic grasp metrics from partial information about a target object. %We refer to Section~\ref{subsec:inference} for the formal definition of $\mathcal{M}^\Phi$.
\end{enumerate}

We provide an implementation of the metric extraction function $\Phi$ for the selected metrics (which we describe in Section~\ref{subsec:metrics}) as a plugin for the GraspIt!~\cite{GraspIt!} simulator, some of which are formulated as linear programming problems which we solve through the CGAL library~\cite{cgal2008computational}. 
Input object models are selected from the Princeton Shape Benchmark~\cite{shilane2004princeton} dataset and assumed to be constituted of homogeneous plastic, and grasps are produced using the model of a Barrett hand~\cite{BarrettHand}. All simulated grasps are evaluated and results are logged into a dataset, preserving both stable and unstable grasps. 
We structure the model $\mathcal{M}^\Phi$ as a classifier that filters stable grasps only and a regressor that estimates the metric vector $\phi$ of stable grasps from the available partial information.

\subsection{Data collection}

\label{subsec:data}
We collected a dataset of grasp and use hypotheses to compute metrics for learning purposes. We sample pregrasps and use directions with uniform distribution in their domain and then simulate the grasp and determine the exact use location on the mesh of an object from the Princeton Shape Benchmark~\cite{shilane2004princeton}. The grasp is simulated with GraspIt!~\cite{GraspIt!} on a Barrett hand~\cite{BarrettHand} with the policy in Figure~\ref{fig:grasp_policy}: from an initial hand position and orientation, the manipulator advances in a straight line until a first contact is made with the object, then the three fingers of the Barrett hand are closed independently until a contact is made or the finger is completely closed. 
 \begin{figure}
    \centering
  \subfloat[\label{fig:barrett_dof:a}]{%
       \includegraphics[width=0.27\linewidth,trim={60 60 60 60},clip]{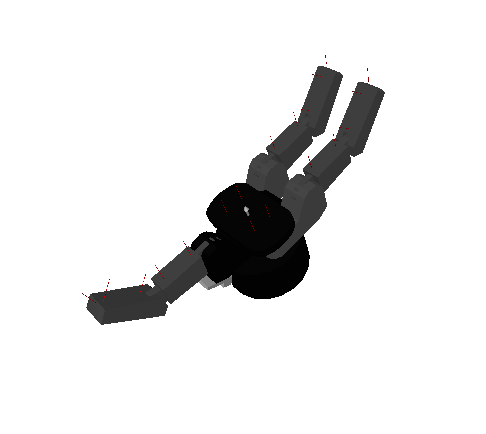}}
    \hfill
  \subfloat[\label{fig:barrett_dof:b}]{%
        \includegraphics[width=0.27\linewidth,trim={60 60 60 60},clip]{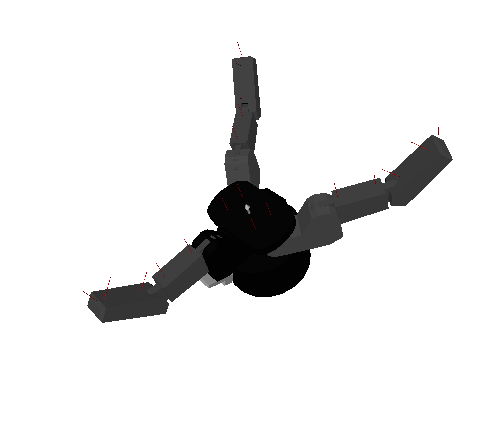}}
    \hfill
  \subfloat[\label{fig:barrett_dof:c}]{%
        \includegraphics[width=0.27\linewidth,trim={60 60 60 60},clip]{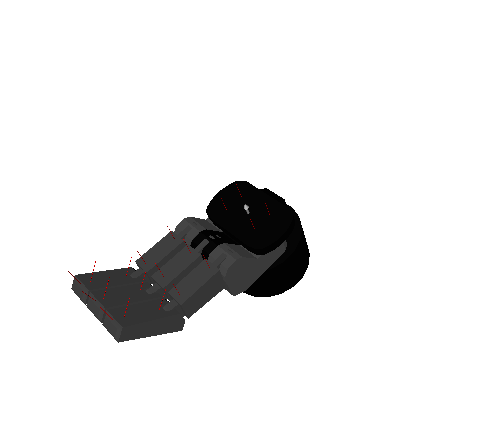}}
  %\caption{The pregrasp degree of freedom of the Barrett hand. (a) DOF set to 0 (b) DOF set to 0.25 (c) DOF set to 1}
  \caption{Pregrasp degree of freedom (a) set to 0, (b) set to 0.25, (c) set to 1}
  \label{fig:barrett_dof} 
\end{figure}

\begin{table}[!t]
\renewcommand{\arraystretch}{0.8}
\caption{Affordance function thresholds}
\label{tab:affordance_function_hyperparameters}
\centering
\begin{tabular}{c|c|c|c}
\hline
\bfseries Mode name & \bfseries $\tau_\rob$ & \bfseries $\tau_\useGeom$ & \bfseries $\tau_\dischargeEfficiency$\\
\hline\hline
Default & 0.3 & 10 & 0.95\\
\hline
No robustness required & $-\infty$ & 10 & 0.95\\
\hline
Extra robustness required & 0.5 & 10 & 0.95\\
\hline
\end{tabular}
%\vspace{-0.2cm}
\end{table}

\subsection{Optimization of Task Grasp Metrics via Simulation}
We consider only one pregrasp degree of freedom for the Barrett hand to encode the angle between the two joint fingers as shown in Figure~\ref{fig:barrett_dof}, thus the domain in which we uniformly randomize the pregrasps is $[-1, 1]^5 \times [0, 1]$: two values in $[-1, 1]$ encode the hand approach direction in normalized spherical coordinates, one value in $[-1, 1]$ encodes the hand rotation around its approach axis, two values in $[-1, 1]$ are the approach offset on the xy-plane relatively to the bounding box of the considered object and one value in $[0, 1]$ is the pregrasp degree of freedom of the Barrett hand; use directions are encoded in normalized spherical coordinates in $[-1, 1]^2$. 
Data collection can be run in parallel on multiple cores and machines, producing millions of data samples each day. Running on 20 cores of an Intel Xeon E5-2630 v4 for a week we could produce a dataset of over 400 million samples of random grasps with metrics, out of which 20 million samples are viable grasps which respect the condition in Section~\ref{subsec:learning}.

\begin{figure}[!t]
    \centering
  \subfloat[\label{fig:joint_pinching:a}]{%
       \includegraphics[width=0.4\linewidth]{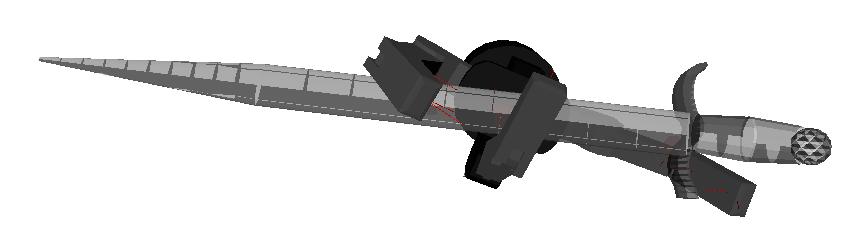}}
    \hfill
  \subfloat[\label{fig:joint_pinching:b}]{%
        \includegraphics[width=0.4\linewidth]{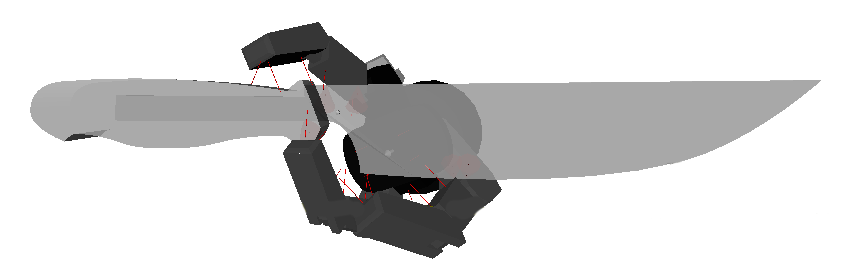}}
    \\
  \subfloat[\label{fig:joint_pinching:c}]{%
        \includegraphics[width=0.4\linewidth]{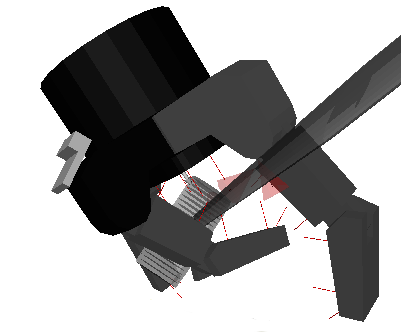}}
    \hfill
  \subfloat[\label{fig:joint_pinching:d}]{%
        \includegraphics[width=0.4\linewidth]{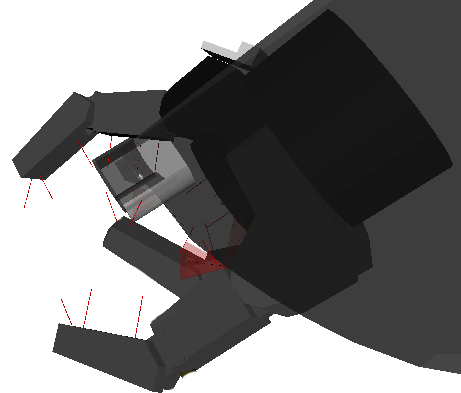}}
  \caption{Optimized grasps from the picking showing the joint pinch strategy. The edge of the blade is pinched between two joints to improve the stability of the grasp; (c) and (d) show the detail of the joint pinch on the blade.}
  \label{fig:joint_pinching} 
\end{figure}

\begin{figure}[!t]
    \centering
  \subfloat[\label{fig:cutting:a}]{%
       \includegraphics[width=0.5\linewidth]{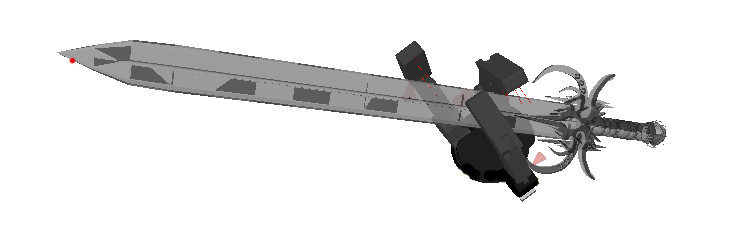}}
    \hfill
  \subfloat[\label{fig:cutting:b}]{%
        \includegraphics[width=0.5\linewidth]{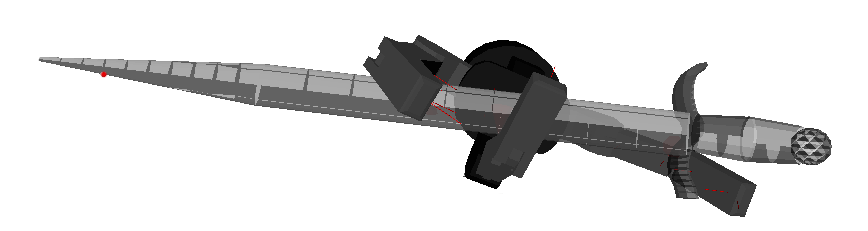}}
    \\
  \subfloat[\label{fig:cutting:c}]{%
        \includegraphics[width=0.5\linewidth]{m724cutting.png}}
    \hfill
  \subfloat[\label{fig:cutting:d}]{%
        \includegraphics[width=0.5\linewidth]{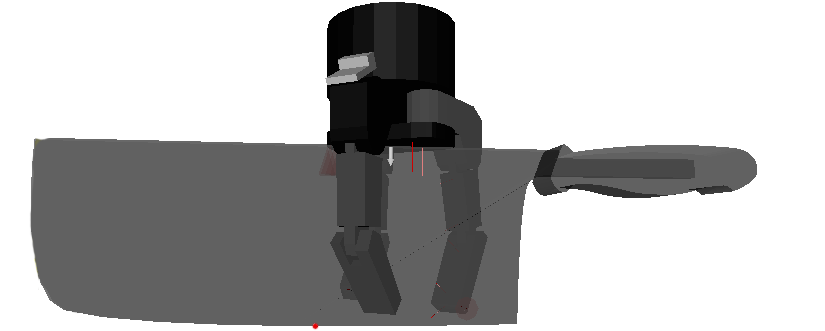}}
  \caption{Optimized grasps from the cutting task showing two different cutting strategies: thin blade tools like (a) and (b) exercise pressure by rotation, wide blade tools like (c) and (d) instead prefer a direct pressure strategy.}
  \label{fig:cutting} 
\end{figure}

\begin{figure}[!t]
    \centering
  \subfloat[\label{fig:picking:a}]{%
       \includegraphics[width=0.37\linewidth]{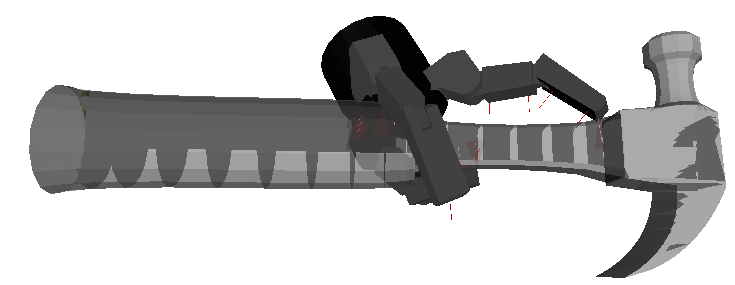}}
%    \hfill
  \subfloat[\label{fig:picking:b}]{%
        \includegraphics[width=0.37\linewidth]{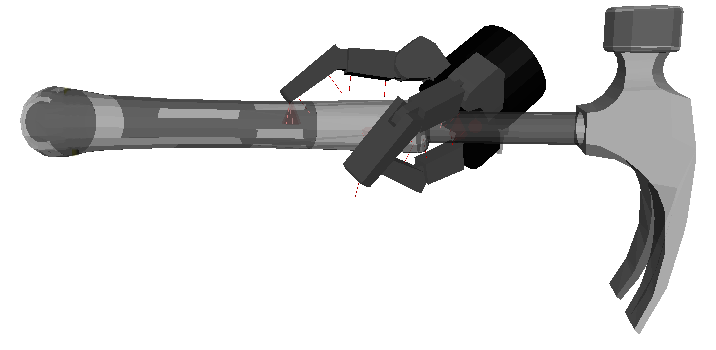}}
    \\
  \subfloat[\label{fig:picking:c}]{%
        \includegraphics[width=0.37\linewidth]{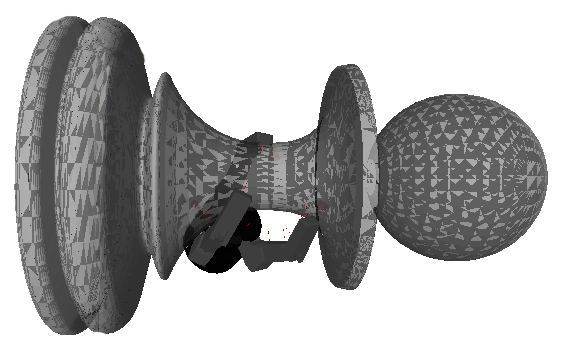}}
%    \hfill
  \subfloat[\label{fig:picking:d}]{%
        \includegraphics[width=0.37\linewidth]{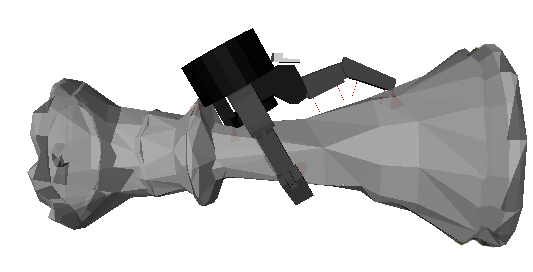}}
  \caption{Optimized grasps from the picking task on sample objects.}
  \label{fig:picking} 
\end{figure}

\begin{figure}[!t]
    \centering
  \subfloat[\label{fig:beating:a}]{%
       \includegraphics[width=0.45\linewidth]{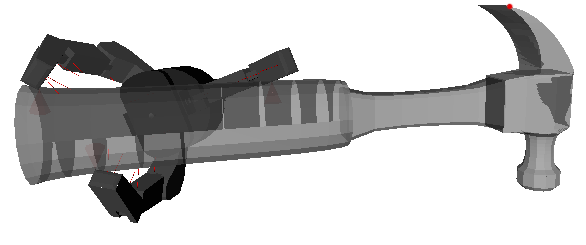}}
    \hfill
  \subfloat[\label{fig:beating:b}]{%
        \includegraphics[width=0.45\linewidth]{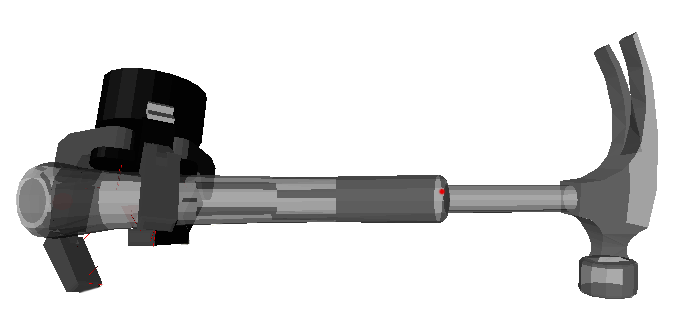}}
    \\
  \subfloat[\label{fig:beating:c}]{%
        \includegraphics[width=0.45\linewidth]{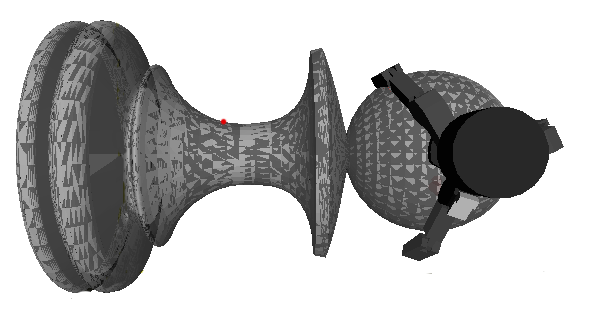}}
    \hfill
  \subfloat[\label{fig:beating:d}]{%
        \includegraphics[width=0.45\linewidth]{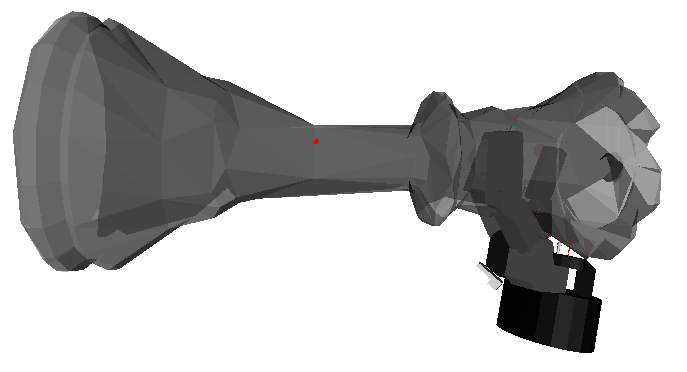}}
  \caption{Optimized grasps from the beating task on sample objects. Notice that the center of mass of hammers in (a) and (b) is on the handle, as the material is assumed homogeneous.}
  \label{fig:beating} 
\end{figure}

To assess the validity of our framework we extract $\argmax_{G, U} \tilde{F}_T(\Phi(O, G, U))$ for our three selected tasks by brute force search on our dataset samples. The parametric thresholds used are the default values reported in Table~\ref{tab:affordance_function_hyperparameters}. 
Sample results of this procedure are shown in Figures~\ref{fig:joint_pinching},~\ref{fig:cutting},~\ref{fig:picking}, and~\ref{fig:beating}. Some of the produced grasps do not appear intuitive, such as the grasps produced for moving blades in Figure~\ref{fig:joint_pinching}, because they exploit features of their physical actuator which are very different from a human hand. In this particular case we observe that the hand achieves a stable grasp by pinching the edge of the blade between the joint of some finger. This pinch provides multiple contacts with very different normals that provide much greater stability of the grasp on a low friction material. The emergence of such solution is very unlikely to happen from human evaluation of grasps, as the human intuition is heavily biased by the human hand with much more fingers, much higher tangential and torsional friction and more susceptible to damage than the metal Barrett hand assumed in our experiments. 

Adaptive behaviours are evident in Figure~\ref{fig:cutting}: with thin blades the hand exercises pressure on the edge by rotation, using the handle as a fulcrum, while with larger blades where such technique is not feasible a direct pressure is preferred.
Picking grasps (Figure~\ref{fig:picking} and~\ref{fig:joint_pinching}) generally wrap around the center of mass as a direct result of the minimization of the total contact forces for holding against gravity.
Beating grasps (Figure~\ref{fig:beating}) display greater variance and generally achieve a stable grasp on the object far from the center of mass to increase the rotational inertia of the object and choose a use location very well aligned with the rotation direction to effectively discharge the rotational energy on the target (the values of $\dischargeEfficiency$ for the optimal grasps are far nearer to 1 than the required values for the selected threshold $\tau_\dischargeEfficiency$). The use location is selected slightly off the center of mass from the opposite side of the hand to balance the beating impulse and produce a torque that contrasts the rotational inertia of the beating movement. 

\begin{figure}
    \centering
  \subfloat[Beating, no robustness required\label{fig:comparison:a}]{%
       \includegraphics[width=0.3\linewidth]{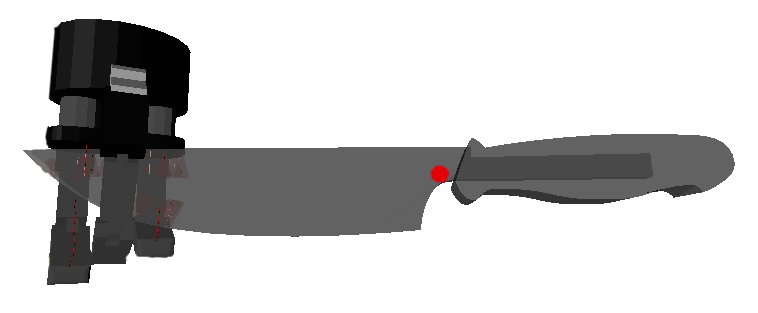}}
    \hfill
  \subfloat[Beating, default\label{fig:comparison:b}]{%
        \includegraphics[width=0.3\linewidth]{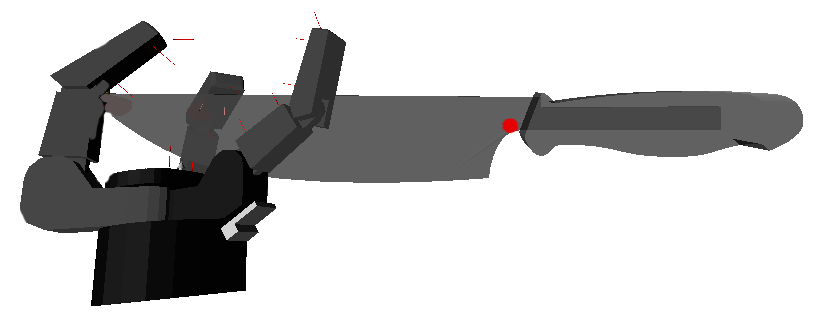}}
    \hfill
  \subfloat[Beating, extra robustness required\label{fig:comparison:c}]{%
        \includegraphics[width=0.3\linewidth]{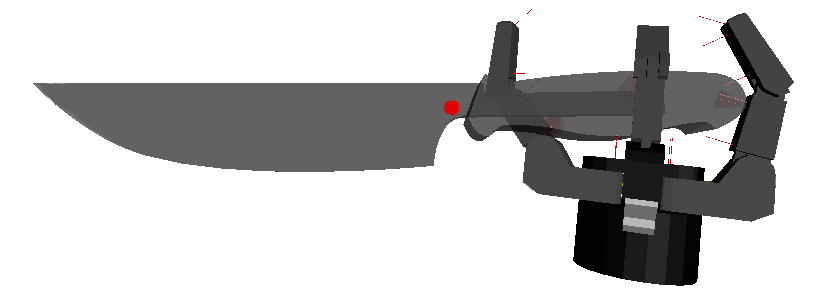}}
    \\
  \subfloat[Cutting, no robustness required\label{fig:comparison:d}]{%
        \includegraphics[width=0.3\linewidth]{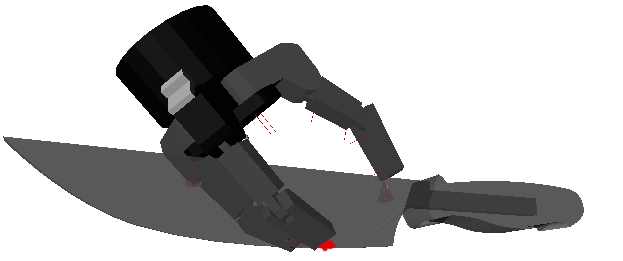}}
    \hfill
  \subfloat[Cutting, default\label{fig:comparison:e}]{%
        \includegraphics[width=0.25\linewidth]{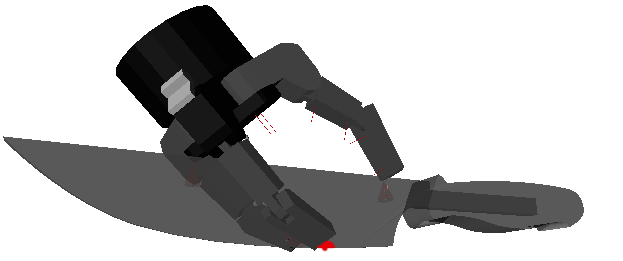}}
    \hfill
  \subfloat[Cutting, extra robustness required\label{fig:comparison:f}]{%
        \includegraphics[width=0.3\linewidth]{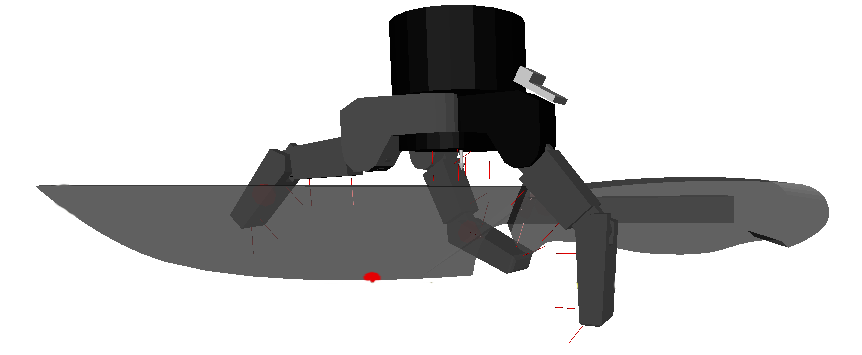}}
  \caption{Affordance function parameter variability. The grasp strategy varies slightly while changing the requirement of the robustness of grasp.}
  \label{fig:comparison} 
\end{figure}

As a further validation test we changed the $\rob$ threshold requirement for the tasks of beating and cutting according to Table~\ref{tab:affordance_function_hyperparameters}. The results of such experiment conducted on the model of a typical kitchen knife are shown in Figure~\ref{fig:comparison}. The produced grasps for beating with a knife display a similar strategy with respect to the ones for more classical objects as seen in Figure~\ref{fig:beating}: the grasp is as far as possible from the center of mass, switching to the handle only when greater robustness is required. The cutting task on the knife shows less variance, producing robust grasps even when not explicitly required by the fitness function.

\subsection{Learning Grasp Metrics from Vision}
\label{subsec:learning}
For the learning phase we defined exactly what a viable grasp should be to generate the ground truth for the classifier network and define the actual dataset for the regressor. We define a viable grasp sample if:
$$
   \rob > \tau_\rob ~~\wedge~~
   \sum_{i=1}^{6} \handEffortHold[i] < \tau_{\handEffortHold} ~~\wedge~~
   \handEffortImpact < \tau_{\handEffortImpact}
$$
where empirically we set $\tau_\rob = 0.15$, $\tau_{\handEffortHold} = 250$, $\tau_{\handEffortImpact} = 100$. 

\begin{figure}
    \centering
    \subfloat[\label{fig:range_image:a}]{%
        \includegraphics[width=0.4\linewidth]{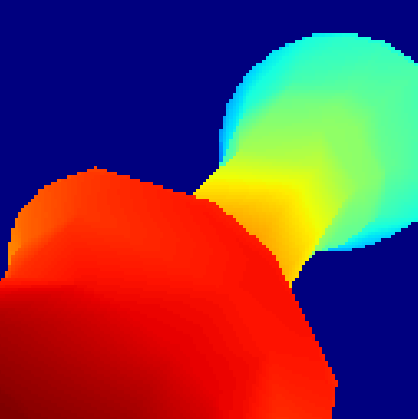}}
    \hfill
  \subfloat[\label{fig:range_image:b}]{%
        \includegraphics[width=0.4\linewidth]{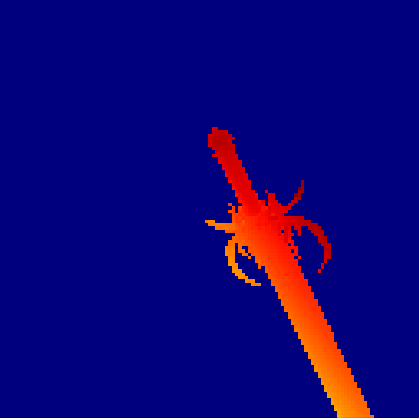}}
  \caption{Synthesized range images with camera-in-hand perspective.}
  \label{fig:range_image} 
\end{figure}

Range images are generated using OpenGL with a perspective projection using common parameters. We take the Kinekt 2 depth camera as a reference with a 70x60° field of view angle: we generate subsampled images of resolution 128x128 using a 60° field of view taken from an object-length distance from the center of the object, as shown in Figure~\ref{fig:range_image}. We formally consider point clouds equivalent to range images as they are generated to hold the same exact information as the input image. We use Open3D~\cite{Zhou2018} to project the synthesized range image to a point cloud which is cleaned from the background. The resulting point cloud is either randomly subsampled of filled with extra points in the origin to match a standard number of points to build batches for efficient learning.
We selected the training, validation and test set from the Princeton Shape Benchmark to propose examples of very different geometries in the training set (1 hammer, 1 screwdriver, 2 bottles, 1 glass, 1 sword, 1 dagger, 1 meat cleaver, 2 ice creams) and to propose similar interesting semantic categories in the validation (1 hammer, 1 bottle, 1 sword, 1 dagger, 1 knife) and test (1 screwdriver, 1 bottle, 1 axe) sets. Validation and test comprise different objects from train but with overlapping semantic categories. 

% Both networks have been trained using dropout~\cite{srivastava2014dropout} layers both in the convolutional and the fully connected segments with drop probability of 0.8 to reduce overfitting. We employed the Adam optimizer~\cite{kingma2014adam} with initial $10^{-3}$ learning rate decaying with inverse proportionality with the epochs and used early stopping with patience of 5. 

\paragraph{The Classifier} filters viable grasps from input data based on the pregrasp and on the input range image with camera-in-hand perspective. As the overall data distribution from our random policy is highly biased towards the negative class (20 times more likely than the positive class), we sample the training data to balance positive and negative samples. 

\paragraph{The Regressor} infers the metric vector $\phi$ for later computation of the affordance function relative to the input sample. In this work we trained regressor networks to infer the value of $\sum_{i=1}^{6} \handEffortHold[i]$ for testing with the picking affordance function on the picking task. Output values are linearly normalized in the interval $[0, 1]$ from the original domain $[0, \tau_{\handEffortHold}]$ granted by the assumption that input values come from the positive class of viable grasps. 

%\subsection{Learning results evaluation}
%\begin{table}[!t]
%\renewcommand{\arraystretch}{0.8}
%\caption{Results on SVM Bag classifier}
%\label{tab:classifier:results}
%\centering
%\begin{tabular}{c|c|c|c|c|c}
%\hline
%\bfseries Target set & \bfseries Accuracy & \bfseries 1-prec. & \bfseries 1-recall & \bfseries 0-prec. & %\bfseries 0-recall\\
%\hline\hline
%%Train & 0.71 & 0.70 & 0.73 & 0.72 & 0.69\\
%Train & 0.72 & 0.71 & 0.75 & 0.74 & 0.69\\
%\hline
%%Validation & 0.65 & 0.66 & 0.62 & 0.64 & 0.69\\
%Validation & 0.66 & 0.66 & 0.65 & 0.65 & 0.67\\
%\hline
%%Test & 0.63 & 0.63 & 0.34 & 0.62 & 0.85\\
%Test & 0.66 & 0.60 & 0.70 & 0.73 & 0.63\\
%\hline
%\end{tabular}
%\end{table}

 \begin{table}[!t]
\renewcommand{\arraystretch}{0.8}
\caption{Results on regressors}
\label{tab:regressor:results}
\centering
\begin{tabular}{c|c|c|c}
\hline
\bfseries Model & \bfseries MSE & \bfseries Comparison accuracy & \bfseries Expected GMS \\
\hline\hline
PointNet Full & 0.050 & 0.63 & 0.63\\
\hline
PointNet Slim & 0.049 & 0.60 & 0.73\\
\hline
CNN & 0.049 & 0.66 & 0.82\\
\hline
\end{tabular}
\end{table}

\begin{figure}
    \centering
        \includegraphics[width=0.65\linewidth,trim={0, 0, 0, 0},clip]{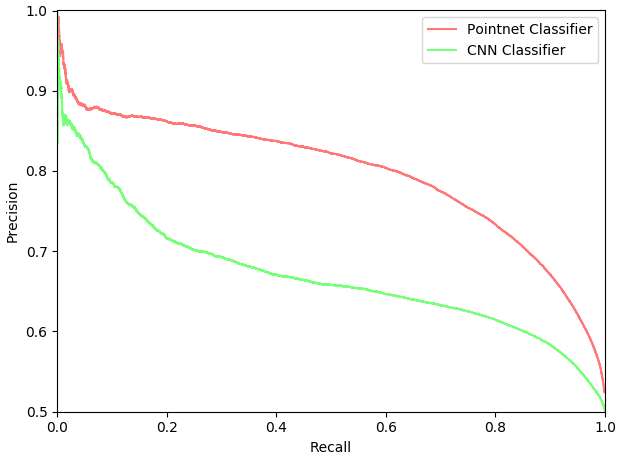}
  \caption{Recall-precision curve for CNN and PointNet classifiers. Curves are drawn on the performance on the test set.}
  \label{fig:recall_precision_curve} 
\end{figure}

The precision-recall curve of the test set for the two trained classifiers are reported in Figure~\ref{fig:recall_precision_curve}. As this classifier model is intended as a filter of good grasp hypotheses, our main metric of interest is the precision of the positive class as this represents the probability that a grasp that passes the filter does really satisfy the expected stability conditions. The recall of the positive class is relevant as well, as it describes the efficiency of the system in missing less good grasp.

 \begin{figure}
    \centering
        \includegraphics[width=0.7\linewidth,trim={20 10 20 30},clip]{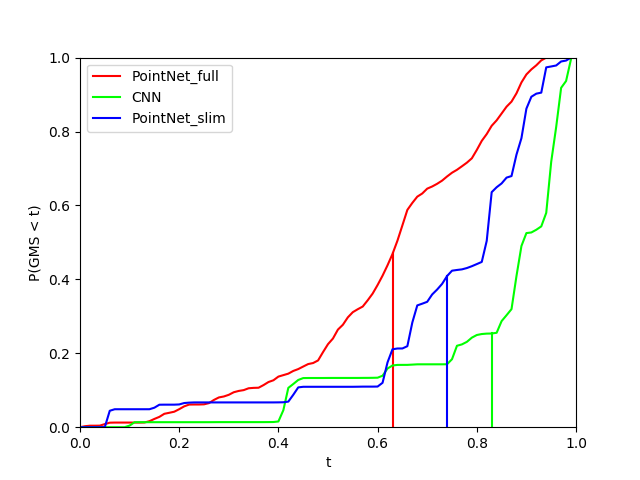}
  \caption{Global minimum score cumulative distributions for the three regressors. The PointNet full regressor is the standard PointNet, the PointNet slim is the standard PointNet with transformation layers fixed to the identity. The vertical lines of each distribution is the expected value for the GMS.}
  \label{fig:GMS_curve} 
\end{figure}

The results on the test set of the trained regressors are in Table~\ref{tab:regressor:results}. The mean squared error, the classical metric used to assess basic regression, gives an overall score of how near the regression goes to real values, but as our goal is optimization, not estimation, we elaborated on two different metrics. As optimization is mainly built by comparisons, we elaborate the comparison accuracy by sampling random couples of samples and measuring the standard accuracy in telling which input sample corresponds to a greater value of the metric. We define the Global Min Score (GMS): let $I_{min}$ be the input sample that minimizes the predicted output $\mathcal{M}^\Phi_R(I_{min})$ over a set $S$ of samples, then the Global Min Score of the model for the set $S$ is the rate of samples that actually have a greater ground truth value than the ground truth value of $I_{min}$. As this score is very sensible to different choices of $S$, we sample random subsets of 10\% of the available samples in $S$ to plot the probability distribution of the value of GMS. Figure~\ref{fig:GMS_curve} shows the cumulative GMS distributions and their respective expected value. This is the most specialized measure of performance of our regressor models as it directly quantifies the relative optimality of the selection of the model relatively to the available choices.

\section{Discussion on Limitations and Future Work}
\label{sec:future_works}
In this work we proposed task dependent metrics to evaluate grasps and use hypotheses in a task-oriented setting. Our framework allows for the automatic evaluation of grasps in a simulation environment and to collect labelled grasp data with minimal human interaction; eliminating the need for human intervention in the grasp labeling process allows both for a widely more scalable data collection and clears the biases that humans have in labeling grasps due to the significant differences between human and robotic hands. 
We showed that we can easily generate millions of labelled grasps on different objects and that roughly hand-designed affordance functions suffice for the emergence of smart and unintuitive techniques for grasping for the exemplified tasks as in Figures~\ref{fig:joint_pinching} and~\ref{fig:cutting}. From this we experimented with convolutional and PointNet~\cite{qi2017pointnet} architectures to learn to infer basic grasp stability and holding hand effort from vision and benchmarked their results, which we consider promising.

There are many directions in which this work can be improved and extended to account for its limitations. The most obvious directions are by validating on more tasks or by accounting for more metrics in simulation such as different materials in compound objects, with different friction coefficients and softness values. This would also provide more realistic locations for the center of mass, which is crucial to determine the optimal grasp in many situations such as for the beating with hammers. As a natural follow up, we foresee a validation step on a real robotic arm to prove the generalization capability from simulation to reality. Indeed, we plan to integrate the learned models with a real Barrett hand and test on similar manipulators (with three or even two fingers) to assess the robustness of performance with inaccurate manipulator models. 
%
%The current hypothesis of camera-in-hand substantially limits the exploration capability of the algorithm that needs to physically move the arm to evaluate different grasp approach directions. To overcome this limitation we plan to extend the current system to consider a voxel grid describing the current geometrical knowledge of the system about the object. Such voxel grid would represent the probability of emptiness of every voxel and can be built from subsequent range images with known techniques~\cite{curless1996volumetric}. This would not only allow to integrate knowledge from multiple range images (not necessarily from the hand) but also to rotate the object representation to face any hypothesized approach direction to search for more appropriate directions with the current integrated knowledge. 
%
An important limitation of the current work is the unstructured definition of the affordance functions $\tilde{F}_T$ which at this stage \textit{define} the tasks themselves for the system. A direction of improvement is towards a different description of tasks from which functions $\tilde{F}_T$ can be extracted or learned. Reference~\cite{PhysicsToolUse2015} extracts a representation of a task in terms of relevant physical quantities from the demonstration of a human choosing a tool and executing the task, such representation can be used to drive the automatic synthesis of the affordance function from a human demonstration. An alternative approach can be the definition of the execution of the task and a performance index of its end effectiveness: this would allow the optimization of the affordance function by reinforcement learning by simulation of the execution of the task itself with no further human intervention.

%\section*{Acknowledgment}
%\tmp{Matt}
%\tmp{somebody for the westworld server}

\bibliographystyle{IEEEtran}
\bibliography{IEEEabrv,references}

\begin{thebibliography}{10}
\providecommand{\url}[1]{#1}
\csname url@samestyle\endcsname
\providecommand{\newblock}{\relax}
\providecommand{\bibinfo}[2]{#2}
\providecommand{\BIBentrySTDinterwordspacing}{\spaceskip=0pt\relax}
\providecommand{\BIBentryALTinterwordstretchfactor}{4}
\providecommand{\BIBentryALTinterwordspacing}{\spaceskip=\fontdimen2\font plus
\BIBentryALTinterwordstretchfactor\fontdimen3\font minus
  \fontdimen4\font\relax}
\providecommand{\BIBforeignlanguage}[2]{{%
\expandafter\ifx\csname l@#1\endcsname\relax
\typeout{** WARNING: IEEEtran.bst: No hyphenation pattern has been}%
\typeout{** loaded for the language `#1'. Using the pattern for}%
\typeout{** the default language instead.}%
\else
\language=\csname l@#1\endcsname
\fi
#2}}
\providecommand{\BIBdecl}{\relax}
\BIBdecl

\bibitem{BlindGraspCorr}
H.~Dang and P.~K. Allen, ``Tactile experience-based robotic grasping,'' in
  \emph{Workshop on Advances in Tactile Sensing and Touch based Human-Robot
  Interaction, HRI}, 2012.

\bibitem{2DBigDataGrasp}
S.~Levine, P.~Pastor, A.~Krizhevsky, J.~Ibarz, and D.~Quillen, ``Learning
  hand-eye coordination for robotic grasping with deep learning and large-scale
  data collection,'' \emph{The International Journal of Robotics Research},
  vol.~37, no. 4-5, pp. 421--436, 2018.

\bibitem{CameraInHand}
D.-J. Kim, R.~Lovelett, and A.~Behal, ``Eye-in-hand stereo visual servoing of
  an assistive robot arm in unstructured environments,'' in \emph{Robotics and
  Automation, 2009. ICRA'09. IEEE International Conference on}.\hskip 1em plus
  0.5em minus 0.4em\relax IEEE, 2009, pp. 2326--2331.

\bibitem{ECVFeatures}
A.~N. Erkan, O.~Kroemer, R.~Detry, Y.~Altun, J.~Piater, and J.~Peters,
  ``Learning probabilistic discriminative models of grasp affordances under
  limited supervision,'' in \emph{Intelligent Robots and Systems (IROS), 2010
  IEEE/RSJ International Conference on}.\hskip 1em plus 0.5em minus 0.4em\relax
  IEEE, 2010, pp. 1586--1591.

\bibitem{AmazonChallenge2017Princeton}
A.~Zeng, S.~Song, K.-T. Yu, E.~Donlon, F.~R. Hogan, M.~Bauza, D.~Ma, O.~Taylor,
  M.~Liu, E.~Romo \emph{et~al.}, ``Robotic pick-and-place of novel objects in
  clutter with multi-affordance grasping and cross-domain image matching,'' in
  \emph{2018 IEEE International Conference on Robotics and Automation
  (ICRA)}.\hskip 1em plus 0.5em minus 0.4em\relax IEEE, 2018, pp. 1--8.

\bibitem{roa2015grasp}
M.~A. Roa and R.~Su{\'a}rez, ``Grasp quality measures: review and
  performance,'' \emph{Autonomous robots}, vol.~38, no.~1, pp. 65--88, 2015.

\bibitem{miller1999examples}
A.~T. Miller and P.~K. Allen, ``Examples of 3d grasp quality computations,'' in
  \emph{Proceedings 1999 IEEE International Conference on Robotics and
  Automation (Cat. No. 99CH36288C)}, vol.~2.\hskip 1em plus 0.5em minus
  0.4em\relax IEEE, 1999, pp. 1240--1246.

\bibitem{BarrettHand}
W.~Townsend, ``Mcb-industrial robot feature article-barrett hand grasper,''
  \emph{Industrial Robot: An International Journal}, vol.~27, no.~3, pp.
  181--188, 2000.

\bibitem{GraspIt!}
A.~T. {Miller} and P.~K. {Allen}, ``Graspit! a versatile simulator for robotic
  grasping,'' \emph{IEEE Robotics Automation Magazine}, vol.~11, no.~4, pp.
  110--122, Dec 2004.

\bibitem{shilane2004princeton}
P.~Shilane, P.~Min, M.~Kazhdan, and T.~Funkhouser, ``The princeton shape
  benchmark,'' in \emph{Proceedings Shape Modeling Applications, 2004.}\hskip
  1em plus 0.5em minus 0.4em\relax IEEE, 2004, pp. 167--178.

\bibitem{gibson1966senses}
J.~Gibson, \emph{The senses considered as perceptual systems}.\hskip 1em plus
  0.5em minus 0.4em\relax Boston: Houghton Mifflin, 1966.

\bibitem{Michaels}
C.~Michaels, ``Affordances: Four points of debate,'' \emph{ECOLOGICAL
  PSYCHOLOGY}, vol.~15, pp. 135--148, 04 2003.

\bibitem{csahin2007afford}
E.~{\c{S}}ahin, M.~{\c{C}}akmak, M.~R. Do{\u{g}}ar, E.~U{\u{g}}ur, and
  G.~{\"U}{\c{c}}oluk, ``To afford or not to afford: A new formalization of
  affordances toward affordance-based robot control,'' \emph{Adaptive
  Behavior}, vol.~15, no.~4, pp. 447--472, 2007.

\bibitem{PreshapesTaskGrasp2007}
M.~Prats, P.~J. Sanz, and A.~P. Del~Pobil, ``Task-oriented grasping using hand
  preshapes and task frames,'' in \emph{Robotics and Automation, 2007 IEEE
  International Conference on}.\hskip 1em plus 0.5em minus 0.4em\relax IEEE,
  2007, pp. 1794--1799.

\bibitem{song2010learning}
D.~Song, K.~Huebner, V.~Kyrki, and D.~Kragic, ``Learning task constraints for
  robot grasping using graphical models,'' in \emph{2010 IEEE/RSJ International
  Conference on Intelligent Robots and Systems}.\hskip 1em plus 0.5em minus
  0.4em\relax IEEE, 2010, pp. 1579--1585.

\bibitem{SemanticGrasping2012}
H.~{Dang} and P.~K. {Allen}, ``Semantic grasping: Planning robotic grasps
  functionally suitable for an object manipulation task,'' in \emph{2012
  IEEE/RSJ International Conference on Intelligent Robots and Systems}, Oct
  2012, pp. 1311--1317.

\bibitem{SemanticGrasping2014}
H.~Dang and P.~K. Allen, ``Semantic grasping: planning task-specific stable
  robotic grasps,'' \emph{Autonomous Robots}, vol.~37, no.~3, pp. 301--316,
  2014.

\bibitem{detry2017task}
R.~Detry, J.~Papon, and L.~Matthies, ``Task-oriented grasping with semantic and
  geometric scene understanding,'' in \emph{2017 IEEE/RSJ International
  Conference on Intelligent Robots and Systems (IROS)}.\hskip 1em plus 0.5em
  minus 0.4em\relax IEEE, 2017, pp. 3266--3273.

\bibitem{Lakani2019}
S.~R. Lakani, A.~J. Rodr{\'\i}guez-S{\'a}nchez, and J.~Piater, ``Towards
  affordance detection for robot manipulation using affordance for parts and
  parts for affordance,'' \emph{Autonomous Robots}, vol.~43, no.~5, pp.
  1155--1172, 2019.

\bibitem{myers2015affordance}
A.~Myers, C.~L. Teo, C.~Ferm{\"u}ller, and Y.~Aloimonos, ``Affordance detection
  of tool parts from geometric features,'' in \emph{2015 IEEE International
  Conference on Robotics and Automation (ICRA)}.\hskip 1em plus 0.5em minus
  0.4em\relax IEEE, 2015, pp. 1374--1381.

\bibitem{ciocarlie2007dexterous}
M.~Ciocarlie, C.~Goldfeder, and P.~Allen, ``Dexterous grasping via eigengrasps:
  A low-dimensional approach to a high-complexity problem,'' in \emph{Robotics:
  Science and Systems Manipulation Workshop-Sensing and Adapting to the Real
  World}.\hskip 1em plus 0.5em minus 0.4em\relax Citeseer, 2007.

\bibitem{qi2017pointnet}
C.~R. Qi, H.~Su, K.~Mo, and L.~J. Guibas, ``Pointnet: Deep learning on point
  sets for 3d classification and segmentation,'' in \emph{Proceedings of the
  IEEE Conference on Computer Vision and Pattern Recognition}, 2017, pp.
  652--660.

\bibitem{cgal2008computational}
C.~Cgal, ``Computational geometry algorithms library,'' 2008.

\bibitem{Zhou2018}
Q.-Y. Zhou, J.~Park, and V.~Koltun, ``{Open3D}: {A} modern library for {3D}
  data processing,'' \emph{arXiv:1801.09847}, 2018.

\bibitem{PhysicsToolUse2015}
Y.~Zhu, Y.~Zhao, and S.~Chun~Zhu, ``Understanding tools: Task-oriented object
  modeling, learning and recognition,'' in \emph{Proceedings of the IEEE
  Conference on Computer Vision and Pattern Recognition}, 2015, pp. 2855--2864.

\end{thebibliography}

\end{document}